\documentclass[a4paper, 10pt, conference]{ieeeconf}      

\IEEEoverridecommandlockouts                              

\title{\LARGE \bf
A Whole-Body Motion Imitation Framework from Human Data for Full-Size Humanoid Robot
}

\author{Zhenghan Chen$^{1,2^*} $, Haodong Zhang$^{1,2^*}$, Dongqi Wang$^{2,3}$, Jiyu Yu$^{1,2}$, Haocheng Xu$^{1,2}$, \\ Yue Wang$^{1}$, Rong Xiong$^{1,2}$ 
\thanks{*The first two authors contributed equally. 1 The State Key Laboratory of Industrial Control and Technology of Zhejiang University. 2 Zhejiang Humanoid Robot Innovation Center. 3 Department of Engineering Mechanics of Zhejiang University. Rong Xiong is the corresponding author. rxiong@zju.edu.cn. This work was supported by the Joint Funds of the National Natural Science Foundation of China (Grant No. U24A20128).}
}

\usepackage{array}
\usepackage{amsmath}
\usepackage{amssymb}
\usepackage{hyperref}
\usepackage{bbm}
\usepackage{bm}
\usepackage{stfloats}
\usepackage{subfigure} 
\usepackage{float} 
\usepackage{multirow}
\usepackage{graphicx}
\usepackage{cleveref}
\usepackage{mathrsfs}
\usepackage{threeparttable}
\begin{document}

\maketitle
\thispagestyle{empty}
\pagestyle{empty}

\begin{abstract}
Motion imitation is a pivotal and effective approach for humanoid robots to achieve a more diverse range of complex and expressive movements, making their performances more human-like. However, the significant differences in kinematics and dynamics between humanoid robots and humans present a major challenge in accurately imitating motion while maintaining balance. In this paper, we propose a novel whole-body motion imitation framework for a full-size humanoid robot. The proposed method employs contact-aware whole-body motion retargeting to mimic human motion and provide initial values for reference trajectories, and the non-linear centroidal model predictive controller ensures the motion accuracy while maintaining balance and overcoming external disturbances in real time. The assistance of the whole-body controller allows for more precise torque control. Experiments have been conducted to imitate a variety of human motions both in simulation and in a real-world humanoid robot. These experiments demonstrate the capability of performing with accuracy and adaptability, which validates the effectiveness of our approach.

\end{abstract}

\section{INTRODUCTION}

Humanoid robot is a type of robot resembling the human body in shape, capable of performing anthropomorphic motion and interacting with human and environments. Anthropomorphic motion, one of the most significant abilities of humanoid robot, is the cornerstone that supports humanoid robot to learn human-specific assignments and improve the performance of anthropomorphic tasks in diverse settings. This enables humanoid robots to imitate human motions, enhancing their capacity to interact with human operators.\cite{c1,c9}. Unfortunately, it is still a challenge for humanoid robot to perform anthropomorphic motions autonomously. 

Different attempts have been made to endow humanoid robots with anthropomorphic motion capabilities. Traditional methods based on the kinematic and dynamic model analysis focus on how to examine the stability criteria and evaluate the balance while tracking the reference trajectories \cite{c14,c16}, but this type of method lacks the consideration about anthropomorphism of motions. Although the end-to-end reinforcement learning (RL) method combined with adversarial training shows unique advantages of synthesizing and executing graceful and life-like behaviors in simulation\cite{c11,c12}. Another significant method for human-like motion imitation is motion retargeting combined with high-performance controllers, which can achieve the highest similarity in motions. Although combining them organically still faces some challenges, this approach has been widely applied and become the possible solution of anthropomorphic motion imitation \cite{c17}.

The first challenge of this combined method is how to deal with the different kinematics and dynamics structures between humans and humanoid robots. Inverse kinematics based on morphing parameter models can improve the accuracy of motions and achieve more natural behaviors, although this may cause the robot to violate its own constraints \cite{c5,c6,c8}. To address this issue, an optimization-based strategy has been introduced, which solves an optimization problem to find the optimal solution while avoiding that the robot approaches its physical constraints limits \cite{c2,c7}. However, these methods required significant time to optimize each motion and may trap in local minimum. Another common strategy is data-driven motion retargeting with the benefits of deep latent variable modeling to overcome these problems \cite{c31}, but it performs inadequately with unfamiliar actions and results in inaccurate movements.

\begin{figure}[t]
      \centering
      \setlength{\abovecaptionskip}{-0.0cm}
      \setlength{\belowcaptionskip}{-0.5cm}
      \includegraphics[width=0.4\textwidth]{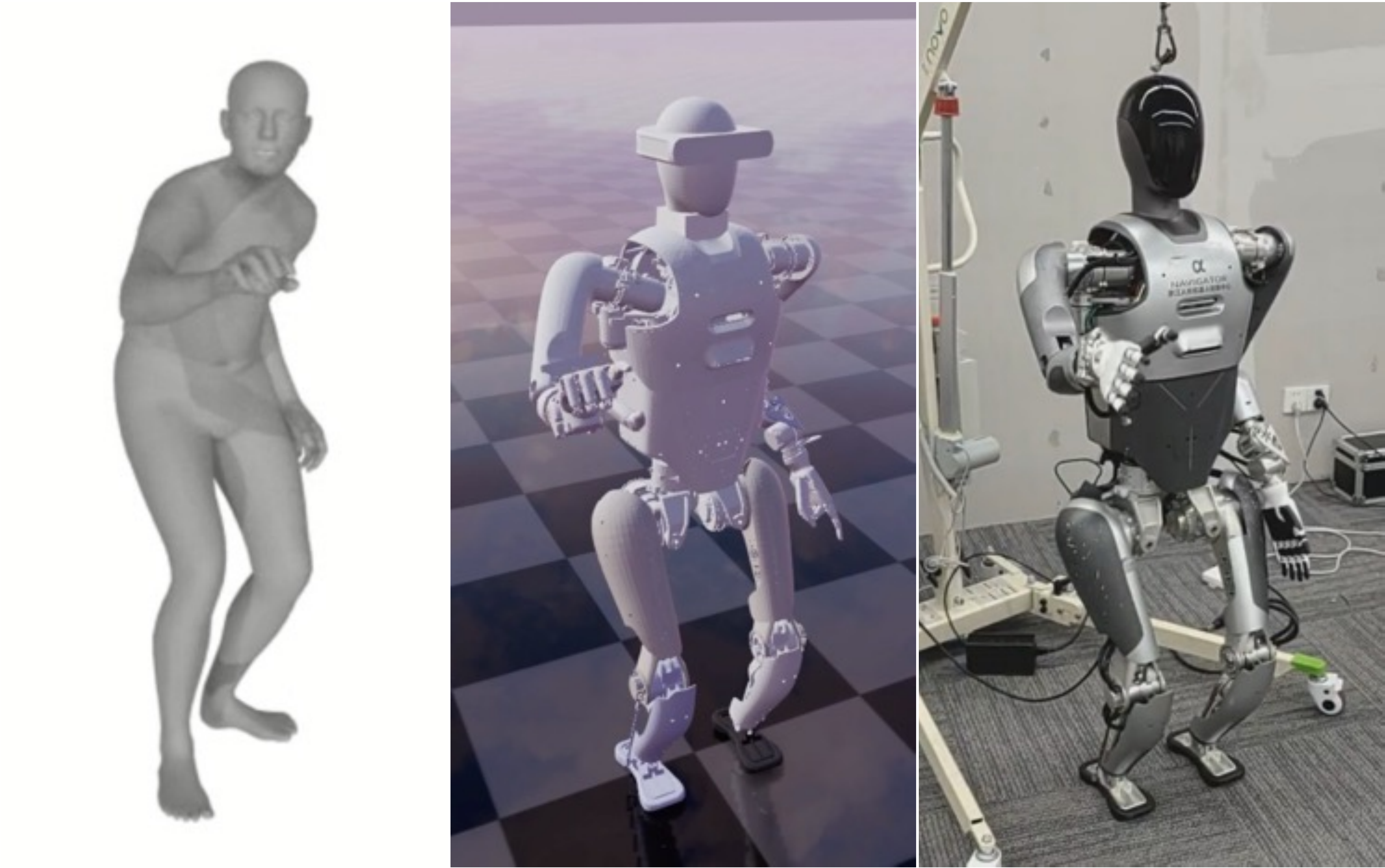}
      \caption{The Screenshot of origin human data, and the motion retargeting result of simulation and robot in real world. The motion retargeting considers the physical constraints of robots and remaps the whole-body motion to humanoid robot.}
      \label{fig::surface}
      \vspace{-18pt}
\end{figure}

When performing anthropomorphic motion imitation, the other significant challenge that the combined method faces: ensuring dynamic balance while imitating the human-like motion. Model-based optimization methods have made trade-offs between model complexity and real-time computation. On one hand, offline optimization considers multi rigid body dynamics to achieve the most accurate model, but it sacrifices real-time computation, making it impossible for real-time robot control \cite{c13,c16}. On the other hand, the simplified model method has been used effectively and is computationally efficient for real-time imitation \cite{c3,c14,c15}. However, if the predetermined assumptions are violated, these methods can fail, resulting in unpredictable motion. Model Predictive Control (MPC) instead \cite{c17,c18}, can improve the ability to respond to sudden reactive events and enhance the expressiveness of motions at the same time, which effectively balances the above two algorithms very well.

Therefore, in this paper, we propose a whole-body motion imitation framework for full-size humanoid robots that combines the advantages of motion retargeting and non-linear MPC. The contact-aware whole-body motion retargeting part can provide good initial values for the subsequent controller. The non-linear MPC based on centroidal dynamics can achieve a real-time solution while ensuring the accuracy and stability of the robot's motion imitation, and ensuring that the robot does not violate any nonlinear dynamic constraints. We present results in both simulation and physical platforms, demonstrating the versatility of our proposed method, which has high tracking accuracy while ensuring robot stability. Our contributions are as follows:
\begin{itemize}
\item propose a unified framework consisting of whole-body motion retargeting and model-based control to enable a full-size humanoid robot to imitate human motion data with anthropomorphism and robustness. 
\item introduce whole-body motion retargeting that mimics diverse and expressive human motion while considering foot contact with the environment to provide a high-quality reference trajectory for motion control.
\item extend non-linear model predictive control (NMPC)\cite{c37} to whole-body humanoid robot to ensure real-time motion tracking accuracy and maintain balance robustly.
\item our method is tested on a real-world full-size humanoid robot that can perform real-time motion imitation of a wide range of whole-body motions, including upper-body movement, one-legged standing, and other quasi-static motions.
\end{itemize}

\begin{figure*}[ht] 
    \vspace{4pt}
    \centering
    \setlength{\abovecaptionskip}{-0.0cm}
    \setlength{\belowcaptionskip}{-0.2cm}
    \includegraphics[width=1.0\textwidth]{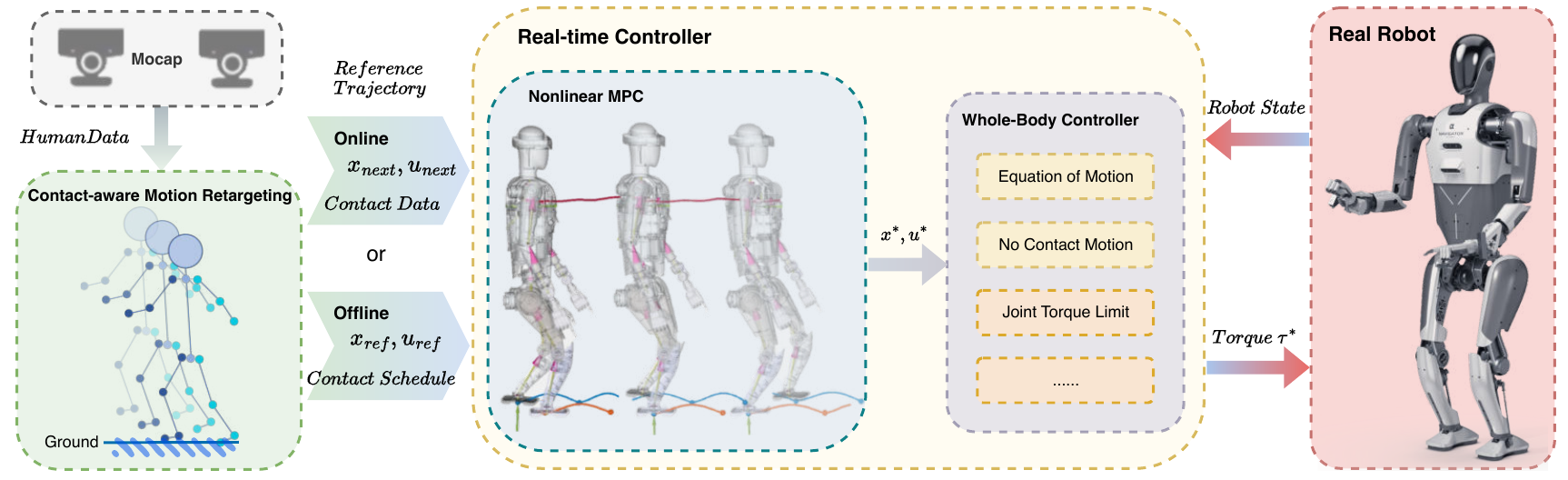}
    \caption{The overall architecture. We first perform whole-body motion retargeting from human motion data to obtain the online reference trajectory at the next moment($x_{next},u_{next}$) with contact data or the whole offline reference trajectory($x_{ref},u_{ref}$) with the contact schedule. Then, non-linear centroidal model predictive control (NMPC) computes the optimal robot state $x^*$ and the optimal control input $u^*$ while robustly resisting disturbances. Finally, the whole-body controller computes the desired torques $\tau^*$, which are executed by the real full-size humanoid robot.}
    \label{fig::framework}
    \vspace{-17pt}
\end{figure*}

\section{RELATED WORKS}
Motion imitation for humanoid robots using various human data has received considerable attention in recent years due to the human-like structure of robots and their adaptability to various tasks. Miura et al. \cite{c19} presented a method for creating human-like locomotion in humanoid robots by remixing motion capture data, aiming to bridge the gap between human movements and robotic capabilities. Pollard et al. \cite{c20}focused on translating human movements into robotic actions while ensuring stability, efficiency, and naturalness of motion. Zhang et al.\cite{c10} proposed a novel kinematic method that utilizes the advantages of both neural networks and optimization to provide a new perspective for mapping upper-body human motions. Darvish et al. \cite{c21} proposed a method based on inverse kinematics that provided real-time anthropomorphic limb motions for humanoid robot joints. Ishiguro et al. \cite{c22} successfully transferred highly dynamic upper-body and leg motions to the humanoid robot JAXON and demonstrated challenging motions. However, these techniques cannot be used for running and jumping motions and lack robustness. Different from previous techniques, we address the whole-body motion retargeting problem with the foot contact information, yielding the foot contact sequence and timings and providing diverse and anthropomorphic reference trajectory for motion control.

Although significant progress has been made in motion retargeting, the other primary challenge in motion imitation is about the controller that ensuring anthropomorphic and stable motion in real time while avoiding falls. Dariush et al. \cite{c14}presents an online task space control retargeting formulation that satisfies balance constraints with a ZMP controller to maintain balance. Yamane et al. \cite{c23} introduced a balance controller based on a simplified robot model and an optimization process to obtain joint torques that enabled a force-controlled robot to maintain balance. However, these methods sacrifice precise models to ensure real-time solving and neglect the re-targeting of torso movements. To satisfy equality and inequality constraints in contact force, Righetti \cite{c24} proposed an inverse dynamic controller that minimizes a combination of linear and quadratic costs. This approach aims to be robust to uncertainties in the dynamic model. Romualdi et al. \cite{c18} proposed an online centroidal MPC formulation for adjusting foot placement, which shows the robustness and potential for more accurate trajectory tracking. As a comparison, our approach integrates real-time non-linear centroidal MPC with contact-aware motion retargeting to guide a humanoid robot imitating anthropomorphic human motion, ensuring accurate and stable motions while maintaining dynamic balance with a better local minimum of the optimization problem.

\section{RETARGETING AND CONTROL}

To address the challenging task of whole-body motion imitation from human data for a full-size humanoid robot, we adopt a three-stage pipeline as depicted in \Cref{fig::framework}. The overall framework first generates a high-quality kinematic reference trajectory with foot contact awareness by whole-body motion retargeting, then incorporates dynamic constraints, maintains balance, and resists environmental disturbances by real-time non-linear centroidal MPC, and finally executes the retargeted motion by whole-body controller.

\subsection{Whole-Body Motion Retargeting}

The objective of whole-body motion retargeting is to mimic the human motion and generate a kinematically feasible reference trajectory for the humanoid robot. The key challenge lies in the morphological differences between human and robot, including link proportions and robot constraints. Given the human motion trajectory, which defined as joint positions $\boldsymbol{P}^{human} \in \mathbb{R}^{T \times n_h \times 3}$, where $n_h$ is the number of human joints and $T$ is the trajectory length, the goal is to obtain the target robot joint angles $\boldsymbol{Q}^{robot} \in \mathbb{R}^{T \times n_j}$, which can be converted to joint positions $\boldsymbol{P}^{robot} \in \mathbb{R}^{T \times n_j \times 3}$ by forward kinematics. Empirically, the robot root state is based on the human pelvis joint and its velocity is scaled according to the corresponding height. The binary foot contact signals are extracted from the original human trajectory by detecting whether the foot height and velocity is close to zero.

Here, we extend our previous proposed dual-arm motion retargeting method \cite{c10} to the whole-body task with consideration of foot contact preservation. 
A graph neural network is designed to transform human motion to the robot joint motion. It first encodes the human motion into latent vectors $\boldsymbol{Z}$ using a graph encoder $f_\theta$ and then decodes into target robot motion with a graph decoder $f_\phi$. The graph network models both the human skeleton and robot structure as graphs. The human motion sequence is represented as a spatial temporal graph, where the node features correspond to the joint positions $\boldsymbol{P}^{human}$ and the edge features correspond to joint offsets $\boldsymbol{S}^{human} \in \mathbb{R}^{n_h \times 3}$. Additionally, the robot motion sequence is also represented as a spatial temporal graph, where the node features are transformed from latent vectors to robot joint angles, and the edge features correspond to robot initial offsets and rotations  $\boldsymbol{S}^{robot} \in \mathbb{R}^{n_j \times 6}$.

\begin{gather}
    \boldsymbol{Z} = f_\theta (\boldsymbol{P}^{human}, \boldsymbol{S}^{human})\\
    \boldsymbol{Q}^{robot} = f_\phi (\boldsymbol{Z}, \boldsymbol{S}^{robot})\\
    \boldsymbol{P}^{robot} = ForwardKinematics(\boldsymbol{Q}^{robot})
\end{gather}

To imitate the human whole-body motion and preserve accurate foot contact, the objective function $L$ is composed of four terms, including the vector similarity loss $L_{vec}$, foot contact loss $L_{foot}$, smoothness loss $L_{smo}$, and self-collision loss $L_{col}$, where $\alpha$, $\beta$, $\gamma$, and $\eta$ are their respective weights.

\begin{equation}
\label{Eq:retarget}
    L = \alpha L_{vec} + \beta L_{foot} + \gamma L_{smo} + \eta L_{col}
\end{equation}

The vector similarity loss $L_{vec}$ maintains motion similarity by aligning normalized vectors of key joints between the human and robot. Let $V$ denote the vector set that includes vectors from hips to knees, from knees to ankles, from shoulders to elbows, and from elbows to wrists.

\begin{equation}
    L_{vec} = \sum_t \sum_{\boldsymbol{v}_j \in V} || \boldsymbol{v}_{j,t}^{human} - \boldsymbol{v}_{j,t}^{robot}||
\end{equation}

The foot contact loss $L_{foot}$ preserves the contact with the ground. Let $C$ denote the foot set in contact with the ground. We encourage the contact schedule of the robot to be consistent with the human. Therefore, we minimize the norm of robot foot height and velocity to enforce foot contact.

\begin{equation}
    L_{foot} = \sum_t \sum_{j \in C} || \boldsymbol{p}_{j,t}^{z,robot}|| + ||\boldsymbol{\dot{p}}_{j,t}^{robot}||
\end{equation}



The smoothness loss $L_{smo}$ encourages temporal motion consistency by penalizing large deviation of joint positions.

\begin{equation}
    L_{smo} = \sum_t || \boldsymbol{q}_{t} - \boldsymbol{q}_{t-1} ||
\end{equation}

The self-collision loss $L_{col}$ penalizes collision between robot links. The robot links are modelled as capsules. Let $D$ be the link pair set that contains self-collision, and $d_{i,j}$ be the distance between link $i$ and $j$.

\begin{equation}
    L_{col} = \sum_t \sum_{(i,j) \in D}e^{-d_{i,j,t}^2}
\end{equation}

\subsection{Nonlinear Model Predictive Control}
The non-linear centroidal MPC is employed to optimize the state and control input trajectories. However, simultaneously optimizing the motion and the contact schedule is much more difficult than optimizing the motion with a predefined sequence of contacts. Therefore, our nonlinear MPC problem assumes that the contact schedule associated with a given locomotion mode (standing, stepping in place, walking) is given by the user or motion retargeting part. At the same time, a desired base pose and velocity trajectory is provided by the motion retargeting module, which provides an initial guess and helps to simplify the optimization problem.

We consider to use a direct multi-shooting method for transforming the continue optimization problem into a finite-dimensional nonlinear program(NLP). To satisfy the requirement of real-time iterations, the sequential quadratic programming(SQP) approach based on KKT optimal conditions combined with quadratic approximation strategy \cite{c4} is critical for speeding up. By the way, at each control instance, only one SQP step is performed according to the real-time iteration scheme \cite{c33}.

Now we define the state vector $\bm{x} \in \mathbb{R}^{6+n_q}$ and control input vector $\bm{u} \in \mathbb{R}^{12+n_j}$ during the optimization:
\begin{equation}
\label{eq:state_vector}
    \bm{x} = [\bm{h}_{com}^T \ \bm{r}^T_B \ \bm{\theta}_B^T \ \bm{q}^T_j]^T, \bm{u} = [\bm{f}^T_{c_{1}} \ \bm{\tau}^T_{c_{1}} \ \bm{f}^T_{c_{2}} \ \bm{\tau}^T_{c_{2}} \  \bm{\dot{q}}_j^T]^T
\end{equation}

The continue nonlinear MPC problem formulation is following:
\begin{equation}
\label{eq:MPC}
    \left\{\begin{array}{l} 
    \min\limits_{\bm{u}(\cdot)} \ \bm{\Phi}(\bm{x}(T)) + \int_0^T \bm{l}(\bm{x}(t),\bm{u}(t),t)dt \\
    subject \ to \ \bm{\dot{x}} = f(\bm{x}(t), \bm{u}(t), t) \\
    \bm{g}^1_{eq}(\bm{x}(t), \bm{u}(t), t) = 0 \\
    \bm{g}^2_{eq}(\bm{x}(t),t) = 0 \\
    \bm{h}_{in}(\bm{x}(t),\bm{u}(t),t) \ge 0 \\
    \bm{x}(0) = \bm{x}_0 \\ 
    \end{array}\right.
\end{equation}
where $\bm{x}(t)$ and $\bm{u}(t)$ are the state and control input at time $t$. $\bm{g}^1_{eq}$ represents equality state-input-related constraints, and $\bm{g}^2_{eq}$ represents equality constraints only related to state. $\bm{h}_{in}$ includes state constraints and state-input-related constraints, such as joint position and velocity limits, friction cone constraints, etc. $\bm{x}_0$ is the current measured state of robot.

\textit{1) Cost:}
The cost function $\bm{l}(\bm{x}(t), \bm{u}(t), t)$ is constructed as a least square cost.
\begin{equation}
    \bm{l}_x = ||\bm{x} - \bm{x_{ref}}||_Q, \bm{l}_u = ||\bm{u} - \bm{u_{ref}}||_R
\end{equation}

\textit{2) Equality Constraints:}
For each foot contact point $c_i$, it should obey the following equality constraints:
\begin{equation}
    \label{eq:equality_constraint}
    \left\{\begin{array}{l} 
    \begin{split}
         &\bm{v}_{c_i} = 0  & \text{if} \ c_i \in \mathscr{C}\\
         &\bm{v}_{c_i} \cdot \bm{\hat}{n} = v^*(t) &  \text{if} \ c_i \in \overline{\mathscr{C}}\\ 
         &\bm{f}_{c_i} = 0, \bm{\tau}_{c_i} = 0 & \text{if} \ c_i \in \overline{\mathscr{C}} \\
    \end{split}
    \end{array}\right.
\end{equation}
where $v_{c_i}$ is the velocity of the potential contact point. $\mathscr{C}$ represents the point is in contact. The Eq.\ref{eq:equality_constraint} imply the following meanings: the foot of stance leg are not allowed to slip with the ground, the foot of swing leg should track a reference trajectory $v^*(t)$ and the forces will be vanished.

\textit{3) Inequality Constraints:} The inequality includes surface friction cone, joint position and torque limits.
Surface friction cone constraint: 
\begin{equation}
\left\{\begin{array}{l} 
    \mu_{s} f^z_{c_i} - \sqrt{{f^x_{c_i}}^2 + {f^y_{c_i}}^2 + \epsilon^2} \ge 0 \\
    Yf_{c_i}^z - |\tau_{c_i}^x| \ge 0 \\
    Xf_{c_i}^z - |\tau_{c_i}^y| \ge 0 \\
    \tau_{max} \ge \tau_{c_i}^z \ge \tau_{min} \\
    \end{array}\right.
\end{equation}
where $\tau_{min} \triangleq -\mu_s(X+Y)f_{c_i}^z + |Yf_{c_i}^x - \mu_s \tau_{c_i}^x| + |Xf_{c_i}^y - \mu_s \tau_{c_i}^y|$, $\tau_{max} \triangleq + \mu_s(X+Y)f_{c_i}^z - |Yf_{c_i}^x + \mu_s \tau_{c_i}^x| - |Xf_{c_i}^y + \mu_s \tau_{c_i}^y|$ \cite{c36}. $X$ and $Y$ is defined as half the length and width of contact surface.



\subsection{Whole-Body Controller}
An optimization-based whole-body controller helps to map the optimized trajectory obtained from the nonlinear MPC to the humanoid robot's joint torque, which considers the full nonlinear rigid body dynamics. While formulating the constrained hierarchical optimization problem, in this work we instead use a single QP to construct the constrained optimization problem, in which the low-priority tasks are embedded in the cost function, and high-priority tasks are treated as constraints. The list of tasks is simuliar to the \cite{c4,c39}.



\section{EXPERIMENTS}

\subsection{Experimental Setup}

We perform experiments on a full-size humanoid robot with 1.5m height, 52kg weight, and 27 degrees of freedom to validate our approach both in simulation and in the real world. And we build the simulation environment in Raisim \cite{c26}. The non-linear MPC runs at an average update rate of 30Hz, with time horizon $T_d=1$s. The MPC problem is solved via OCS2 \cite{c28} and the QP problem in WBC \cite{c39} is solved by qpOASES.

The training dataset for motion retargeting is the large-scale human motion dataset AMASS \cite{c30}. The data is divided into a training set and a test set with a ratio of 8:2. We use Adam optimizer with a learning rate of 1e-4 and a batch size of 16. The training process takes 20 epochs with an NVIDIA GeForce RTX 4090 D GPU and an Intel(R) Core(TM) i9-14900KF CPU. The retargeting network follows the same architecture as \cite{c10}. To adapt to different humanoid robots, the network needs to be retrained.

     

\begin{figure}[ht]
     \hspace{-10pt}   
     \vspace{-10pt}
     \setlength{\abovecaptionskip}{0.0cm}
     \setlength{\belowcaptionskip}{-0.2cm}
     
    \subfigure[High-level yoga balance with contact sequence]{
    \begin{minipage}[c]{0.98\linewidth} 
        \centering
        \includegraphics[width=1.0\linewidth]{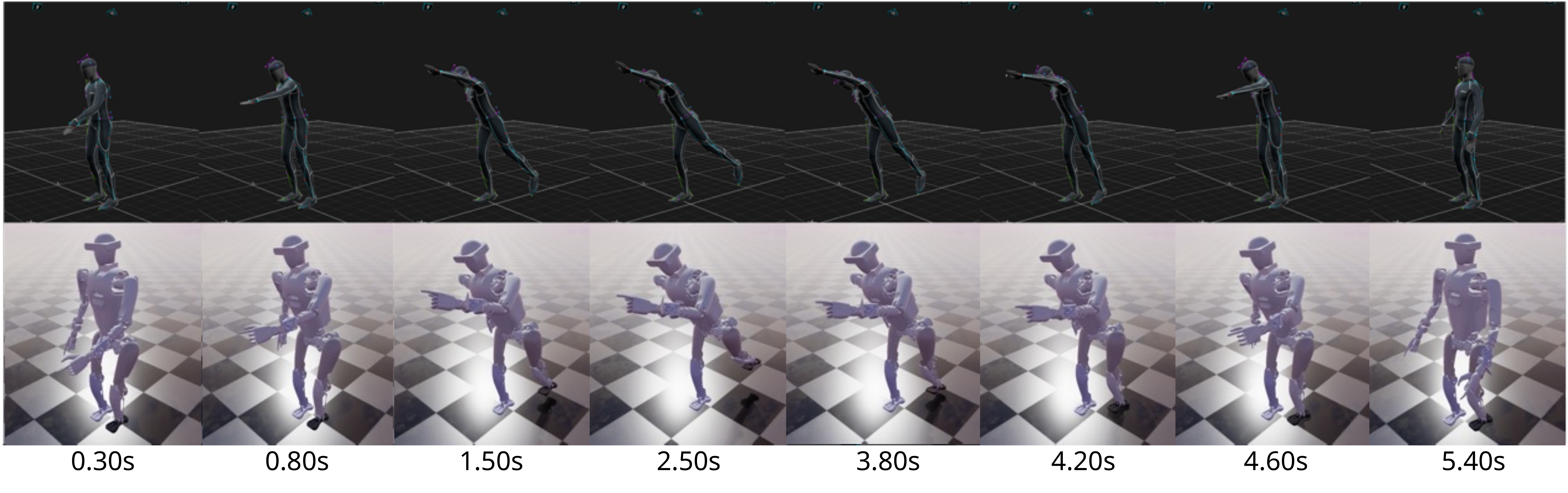}
        \vspace{-12pt}  
    \end{minipage}
    }
    \hspace{-30pt} 
    \vspace{0pt}
    \subfigure[High-dynamic stride walking with contact sequence]{
    \begin{minipage}[c]{0.98\linewidth} 
        \vspace{-5pt}
        \centering
        \includegraphics[width=1.0\linewidth]{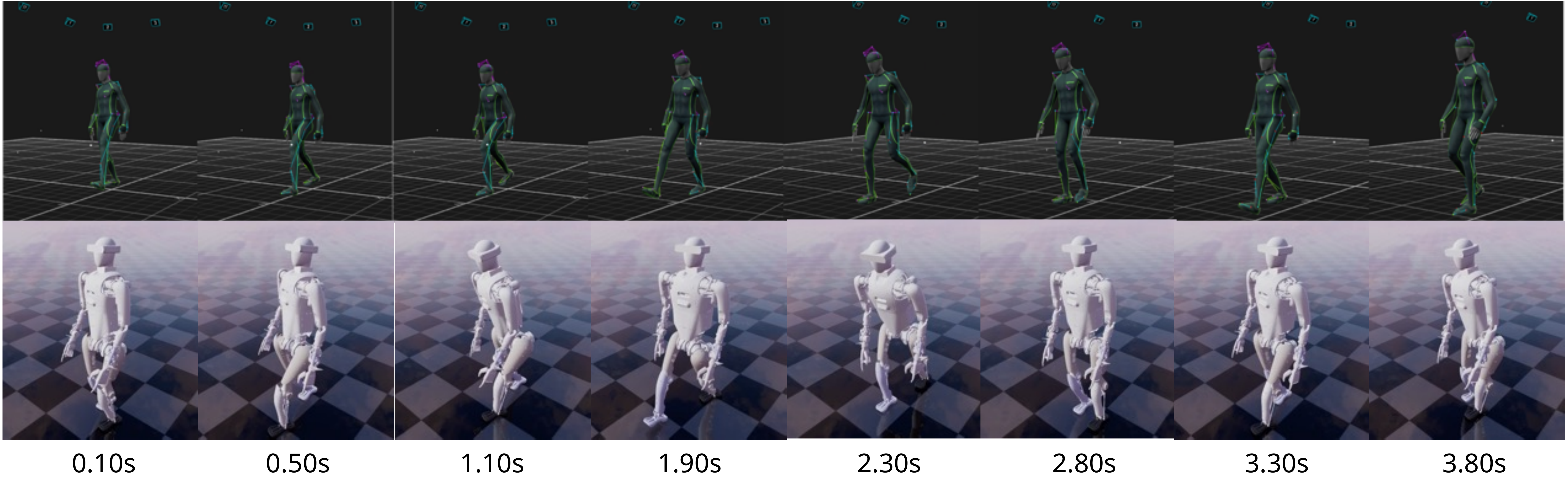}
        \vspace{-12pt}  
    \end{minipage}
    }
    \caption{Human motions that are applied to robot by contact-aware motion retargeting.}
    \label{fig::human_motion}
    \vspace{-14pt}
\end{figure}

\subsection{Case Study}
\noindent\textbf{Offline Motion Retargeting.} We first test our method with pre-recorded human motion data. As shown in the \Cref{fig::human_motion}, our motion retargeting successfully considers the contact sequence between human body and ground contact, and maps it to the robot, successfully performing high-level yoga balance and high-dynamic stride walking motions. 

\begin{figure}[h]
      \vspace{-5pt}
      \setlength{\abovecaptionskip}{-0.1cm}
      \setlength{\belowcaptionskip}{-0.2cm}
      \centering
      \includegraphics[width=0.25\textwidth]{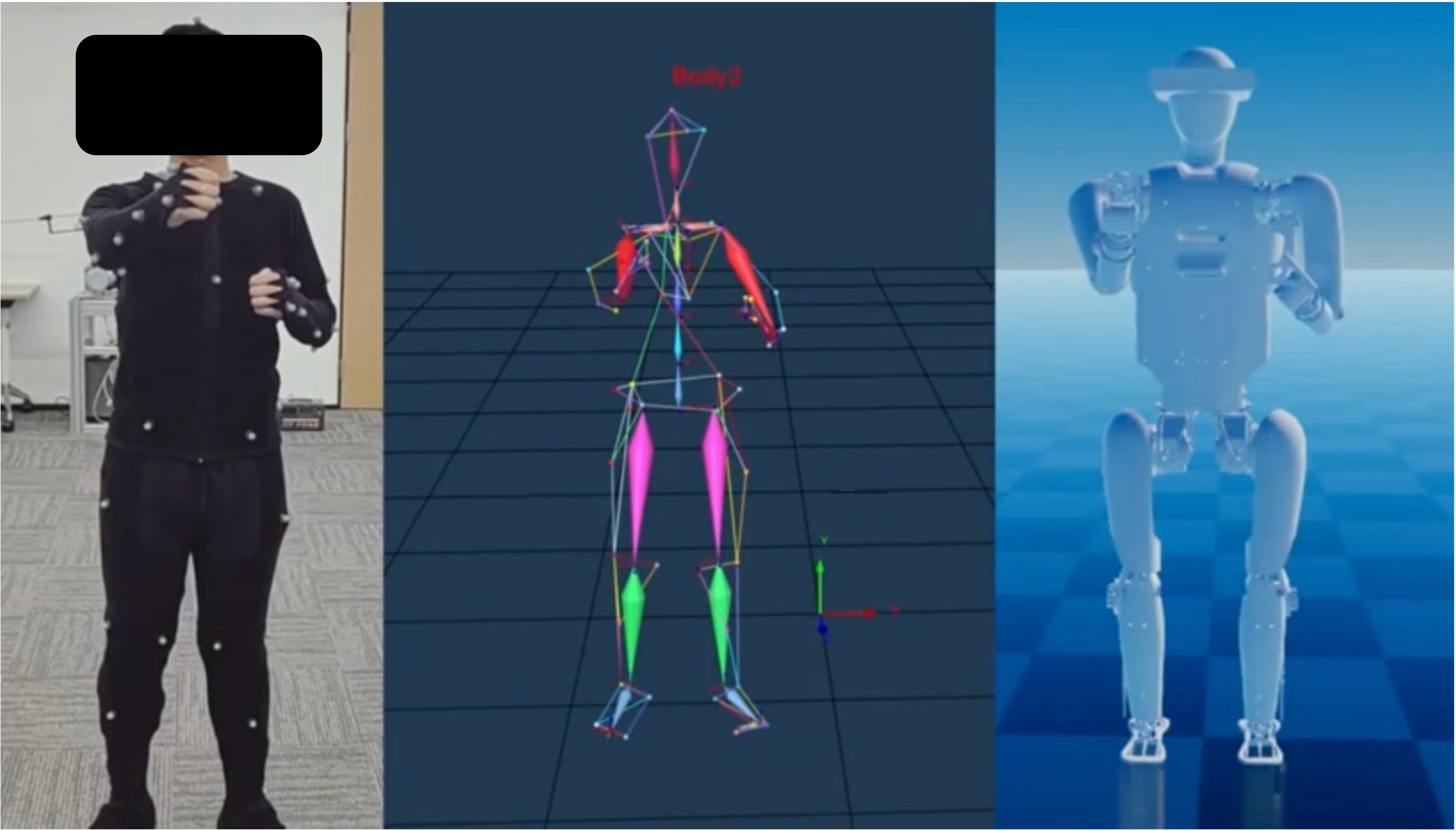}
      \caption{The Screenshot of online motion retargeting.}
      \label{fig::online_retargeting}
      \vspace{-15pt}
   \end{figure}

\noindent\textbf{Online Motion Retargeting.} To achieve real-time online control, the motion retargeting module will generate the reference trajectory for the subsequent moment instead of the whole offline trajectory. The retargeting network generates the robot reference trajectory using a single forward propagation. And linear interpolation will be performed by non-linear MPC module between the current state and the reference trajectory to fill the MPC horizon. As shown in \Cref{fig::online_retargeting}, we can perform real-time online mapping at a frequency of 30Hz, and the average time of each module of the algorithm is shown in \Cref{table_cost_time}.

\begin{table}[h]
    \vspace{-5pt}
    \setlength{\abovecaptionskip}{-0.0cm}
    \setlength{\belowcaptionskip}{-0.2cm}
    \caption{COMPUTATION TIME EVALUATION}
    \label{table_cost_time}
    \centering
    \begin{tabular}{ c c c } 
    \hline
    \textbf{Algorithm} & \textbf{Average Time[ms]} & \textbf{Standard Deviation} \\
    \hline
    Motion Retargeting & 11.34 & 3.6907 \\  
    \hline
    Nonlinear MPC & 23.72 & 1.7076 \\  
    \hline
    Whole-Body Control & 0.31 & 0.5644 \\   
    \hline
    \end{tabular}
    \vspace{-8pt}
\end{table}

\begin{figure}[ht]
  \vspace{-5pt}
  \setlength{\abovecaptionskip}{-0.2cm}
  \setlength{\belowcaptionskip}{0.2cm}
  \centering
  \includegraphics[width=0.47\textwidth]{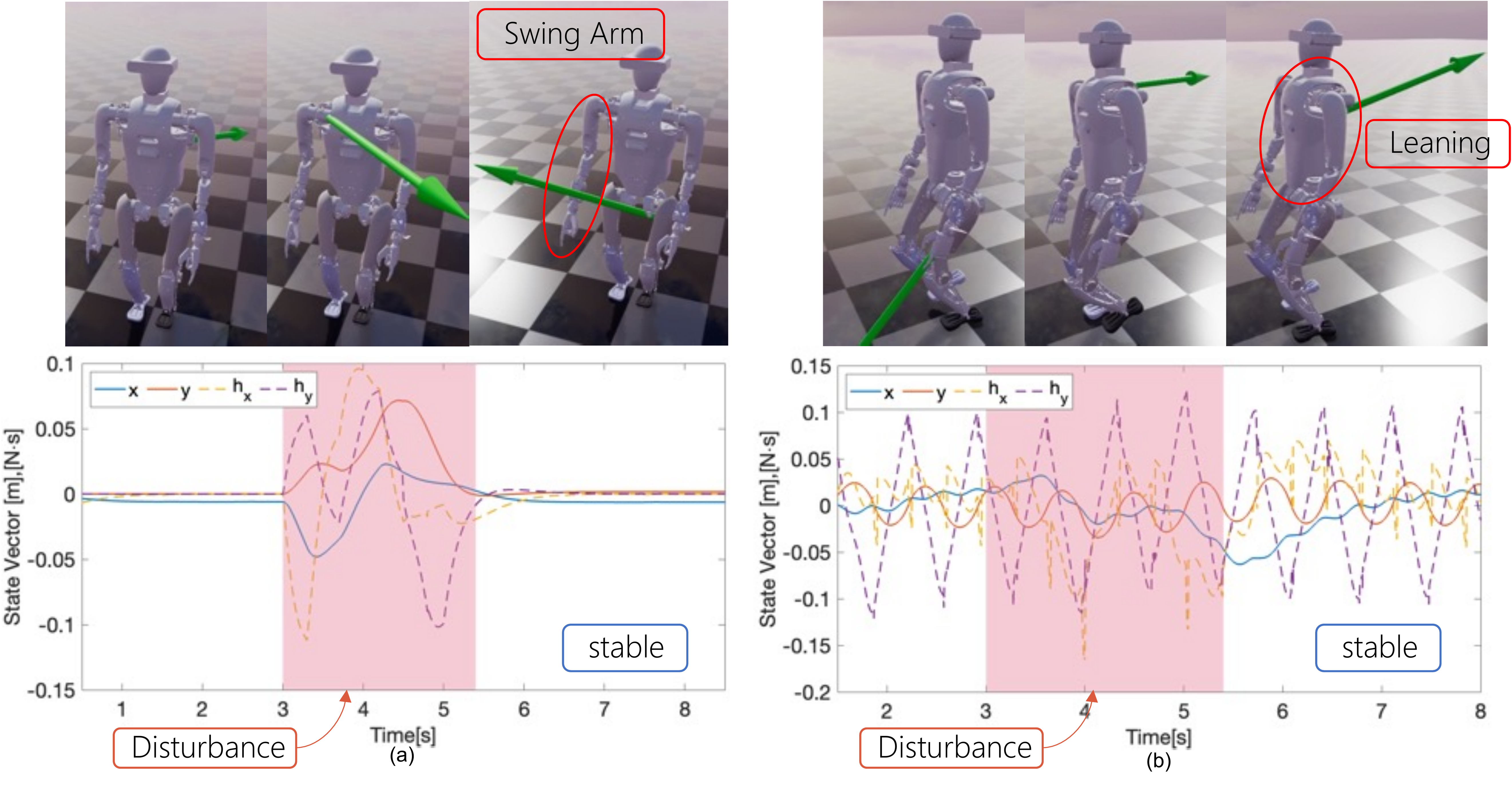}
  \caption{Simulation results that (a) robot is keeping stance under a disturbance of 140 Ns applied in forward and side direction with the magnitude varies within the range of 70N during the marked time of 2.0s. (b) shows that robot is walking under a disturbance of 150 Ns with the magnitude varies within the range of 50N during the marked time of 2.5s.}
  \label{fig::stance_walk_push}
  \vspace{-10pt}
\end{figure}

\noindent\textbf{Push Recovery.}
\Cref{fig::stance_walk_push} demonstrates how the humanoid robot overcomes disturbances while keeping stance or walking with proposed MPC controller. The top of pictures show the reaction of MPC controller to the disturbances, and the robot tries to swing its arms for maintaining balance. The bottom plot shows the change of base position $x,y$ and base momentum $h_x,h_y$. Specifically, our non-linear MPC module is able to maintain stable in several steps while robustly resist the external disturbance, which prove the efficiency of our method.

\noindent\textbf{Generalization to Another Humanoid Robot.} 
The details of experiment about the another humanoid robot are shown in the videos \cite{c38}.

\subsection{Comparative Study}
To thoroughly evaluate the performance of our approach, we compare the trajectory tracking results in terms of success rate and tracking error with baseline methods. We decide to consider linear MPC approach that uses Single Rigid Body Dynamics(SRBD) as a prediction model \cite{c17} as our baseline. And ablation study of using non-linear MPC module also veriﬁes the effectiveness of our approach, that means the results of motion retargeting will be directly given to WBC \cite{c21}. \Cref{table_compare} shows the success rate of our method, the baseline and the ablation study. Both motions place high demands on the robot's balance and joint tracking capabilities.(The COM and joint tracking accuracy of yoga and stride walking plots are shown in the video.)

\begin{table}[h]
    \vspace{-5pt}
    \setlength{\abovecaptionskip}{-0.0cm}
    \setlength{\belowcaptionskip}{-0.2cm}
    \caption{QUANTITATIVE COMPARISON OF DIFFERENT METHODS}
    \centering
    \begin{threeparttable}
    \resizebox{0.47\textwidth}{!}{
        \begin{tabular}{cccccc}
            \hline
            \multirow{2}{*}{\textbf{Motion}} & 
            \multirow{2}{*}{\textbf{Algorithm}} & 
            \multirow{2}{*}{\begin{tabular}[c]{@{}c@{}}\textbf{Success} \\ \textbf{Rate}\end{tabular}} & 
            \multicolumn{3}{c}{\textbf{Average RMSE}} \\ 
            \cline{4-6}
            & & & \multicolumn{1}{c}{\textbf{COM{[}m{]}}} & 
            \multicolumn{1}{c}{\textbf{Leg{[}rad{]}}} & 
            \multicolumn{1}{c}{\textbf{Arm{[}rad{]}}} \\ 
            \hline
            Wave & 
            \begin{tabular}[c]{@{}c@{}c@{}}\textbf{Ours}\\ LMPC+WBC \cite{c17}\\ WBC \cite{c21} \end{tabular} & 
            \begin{tabular}[c]{@{}c@{}c@{}}\textbf{10/10}\\ 10/10\\ 8/10\end{tabular} & 
            \begin{tabular}[c]{@{}c@{}c@{}}\textbf{0.0103}\\ 0.0144\\ 0.0473\end{tabular} & 
            \begin{tabular}[c]{@{}c@{}c@{}}\textbf{0.0535}\\ 0.0643\\ 0.1140\end{tabular} & 
            \begin{tabular}[c]{@{}c@{}c@{}}\textbf{0.0089}\\ 0.0194\\ 0.0882\end{tabular} \\ 
            \hline
            Yoga & 
            \begin{tabular}[c]{@{}c@{}c@{}}\textbf{Ours}\\ LMPC+WBC \cite{c17}\\ WBC \cite{c21} \end{tabular} & 
            \begin{tabular}[c]{@{}c@{}c@{}}\textbf{10/10}\\ 5/10\\ 2/10\end{tabular} & 
            \begin{tabular}[c]{@{}c@{}c@{}}\textbf{0.0170}\\ 0.0400\\ 0.0747\end{tabular} & 
            \begin{tabular}[c]{@{}c@{}c@{}}\textbf{0.1086}\\ 0.1222\\ 0.1893\end{tabular} & 
            \begin{tabular}[c]{@{}c@{}c@{}}\textbf{0.0154}\\ 0.0332\\ 0.1341\end{tabular} \\ 
            \hline
            Stride & 
            \begin{tabular}[c]{@{}c@{}c@{}}\textbf{Ours}\\ LMPC+WBC \cite{c17}\\ WBC \cite{c21} \end{tabular} & 
            \begin{tabular}[c]{@{}c@{}c@{}}\textbf{10/10}\\ 7/10\\ -\end{tabular} & 
            \begin{tabular}[c]{@{}c@{}c@{}}\textbf{0.0472}\\ 0.0598\\ -\end{tabular} & 
            \begin{tabular}[c]{@{}c@{}c@{}}\textbf{0.1299}\\ 0.1901\\ -\end{tabular} & 
            \begin{tabular}[c]{@{}c@{}c@{}}\textbf{0.0359}\\ 0.0405\\ -\end{tabular} \\ 
            \hline
        \end{tabular}
    }
    \begin{tablenotes}
        \item[*] LMPC+WBC \cite{c17} denotes that Linear MPC combined with WBC.
        \item[-] means it failed to complete motion 
    \end{tablenotes}
    \label{table_compare}
    \end{threeparttable}
    \vspace{-10pt}
\end{table}

Notice that compared to linear MPC method, our proposed approach considers a more precise dynamic model and reduces the violation of nonlinear constraints, while slightly improve the tracking accuracy and the success rate of high dynamic motions. Considering ablation study, the non-linear MPC module synthesizes the motion stability within the prediction horizon and enhance the ability to respond to sudden reactive events, which improves both accuracy and robustness.

Since the leg joints are more crucial for maintaining the balance of the humanoid robot, and given the limited dynamic model information in the retargeting process, MPC will control the leg joints to adopt strategies that slightly deviate from the retargeting results in order to best satisfy the robot's centroidal dynamics equation without violating nonlinear constraints. The tracking performance of the arm joints exceeds that of the leg joints. 


\subsection{Real World Experiment}
\begin{figure}[ht]
      \vspace{-10pt}
      \setlength{\abovecaptionskip}{-0.3cm}
      \setlength{\belowcaptionskip}{-0.2cm}
      \centering
      \includegraphics[width=0.4\textwidth]{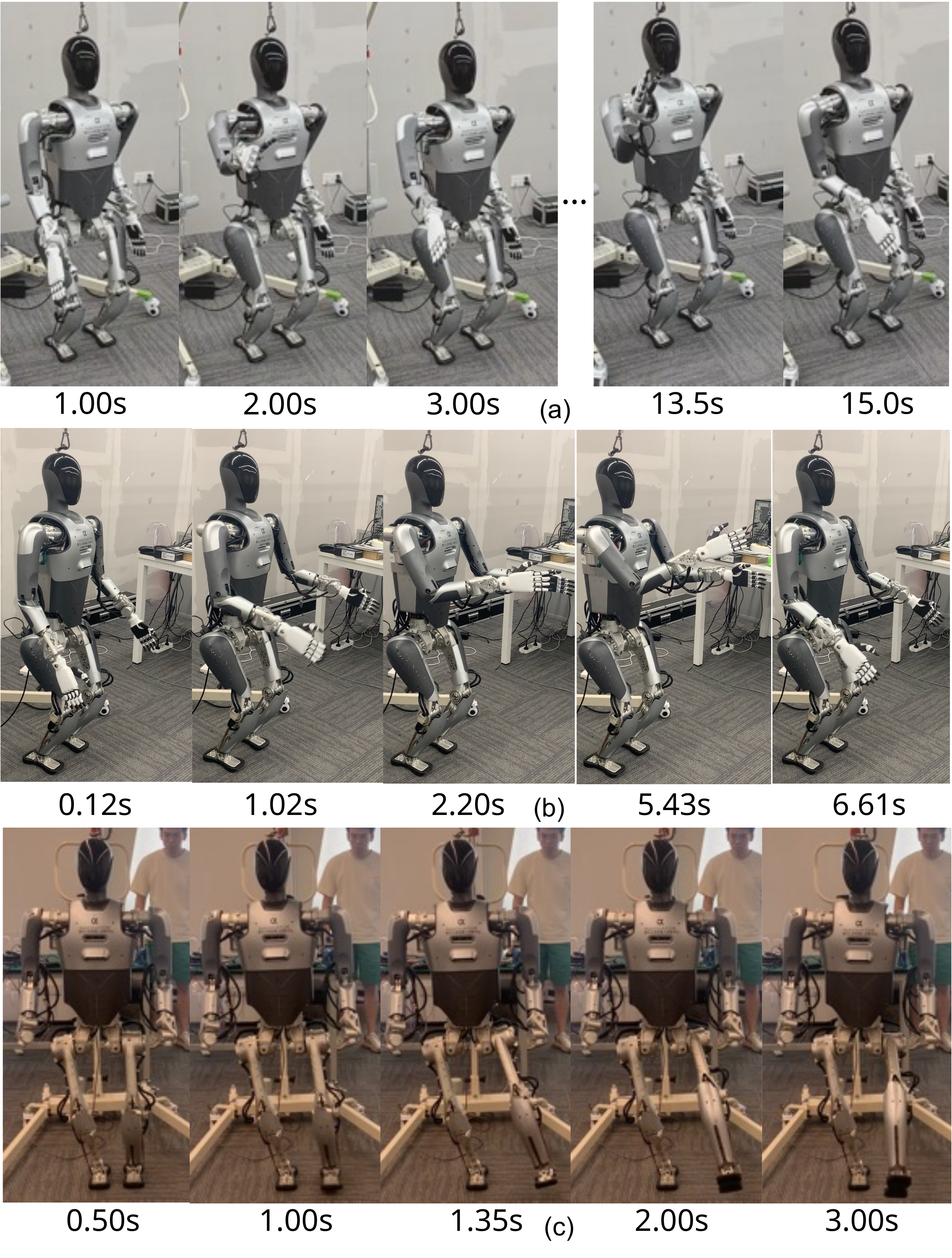}
      \caption{Screenshots of motion imitation. (a) is the upper-body motion retargeting results both in simulation and real world. (b) shows the motion about hand raising. (c) is the task of standing on one leg. The rope in the screenshots is only for protection and will not exert any force on the robot, more details are in videos \cite{c38}. }
      \label{fig::real_robot}
      \vspace{-15pt}
\end{figure}

Our approach was testified in the humanoid robot hardware together with the state estimator. We had verified the upper-body motion while robot in stance state keeping balance, and also completed the task of standing on one leg. \Cref{fig::real_robot} shows the screenshots of the real world experiments, and for each motion we tested five times without failure. The robot employing our strategy is also capable of performing simple walking tasks, shown in the videos \cite{c38}. 

\section{CONCLUSIONS}
In this paper, we propose a novel whole-body motion imitation framework that combines motion retargeting and model-based control for a full-size humanoid robot. The whole-body motion retargeting effectively imitates human motion and captures foot contact, while the model-based controller robustly follows the reference trajectory and adapts to external disturbances. The experimental results show that our method is capable of imitating diverse range of human motions with expressiveness and robustness. The limitation of the proposed method is that it only imitates body motion without dexterous hands, and lacks the consideration about multi-contact maneuvers. Future work will consider extending our method to take into account dexterous hands and multi-contact tasks to achieve more versatile motions. 

\addtolength{\textheight}{0.2cm}   








\enlargethispage{\baselineskip}

\end{document}